\documentclass[10pt,twocolumn,letterpaper]{article}

\usepackage{cvpr}              

\usepackage{graphicx}
\usepackage{amsmath}
\usepackage{amssymb}
\usepackage{amsfonts}
\usepackage{booktabs}
\usepackage{multirow}

\usepackage[pagebackref,breaklinks,colorlinks]{hyperref}

\usepackage[capitalize]{cleveref}
\crefname{section}{Sec.}{Secs.}
\Crefname{section}{Section}{Sections}
\Crefname{table}{Table}{Tables}
\crefname{table}{Tab.}{Tabs.}
\def\cvprPaperID{6} 
\def\confName{CVPR}
\def\confYear{2022}
\begin{document}

\title{WG-VITON: Wearing-Guide Virtual Try-On for Top and Bottom Clothes}

\author{
		Soonchan Park$\textsuperscript{1,2}$ $\hspace{3cm}$ Jinah Park $\textsuperscript{2}$\\
	    $\textsuperscript{1}$ Electronics and Telecommunications Research Institute \\ $\textsuperscript{2}$ Korea Advanced Institute of Science and Technology \\
	    Republic of Korea \\
	{\tt\small parksc@etri.re.kr $\hspace{0.5cm}$ jinahpark@kaist.ac.kr}
}

\maketitle
\begin{abstract}
   Studies of virtual try-on (VITON) have been shown their effectiveness in utilizing the generative neural network for virtually exploring fashion products, and some of recent researches of VITON attempted to synthesize human image wearing given multiple types of garments (e.g., top and bottom clothes). However, when replacing the top and bottom clothes of the target human, numerous wearing styles are possible with a certain combination of the clothes. In this paper, we address the problem of variation in wearing style when simultaneously replacing the top and bottom clothes of the model. We introduce Wearing-Guide VITON (i.e., WG-VITON) which utilizes an additional input binary mask to control the wearing styles of the generated image. Our experiments show that WG-VITON effectively generates an image of the model wearing given top and bottom clothes, and create complicated wearing styles such as partly tucking in the top to the bottom.
\end{abstract}

\section{Introduction}
\label{sec:intro}
Virtual try-on (i.e., VITON) is a task to synthesize an image of a fitting model wearing target garments while maintaining other characteristics of the model such as his/her identity and pose. Various studies have been proposed to improve the quality of the synthesized images wearing given top clothes \cite{han2018viton,wang2018cpvton,minar2020cpvtonplus,yang2020ACGPN,jandial2020sievenet,ge2021parserfree,choi2021vitonhd, ge2021disentangled} based on available dataset such as \cite{han2018viton}. Then, VITON technology was extended for applying multiple types of garment \cite{neuberger2020oviton,li2021ovnet,Cui_2021_Dior}. However, when a person wearing multiple types of the products (e.g., top and bottom clothes), there are diverse wearing styles depending on his/her individual taste. For example, the person can tuck in the top to the bottom, tuck a part of the top to the bottom, or let the top loosely over the bottom. Such \textit{wearing styles} is one of the most important aspects to decide which and how to wear clothes. 

\begin{figure}[t]
	\centering
	\includegraphics[width=7.0cm]{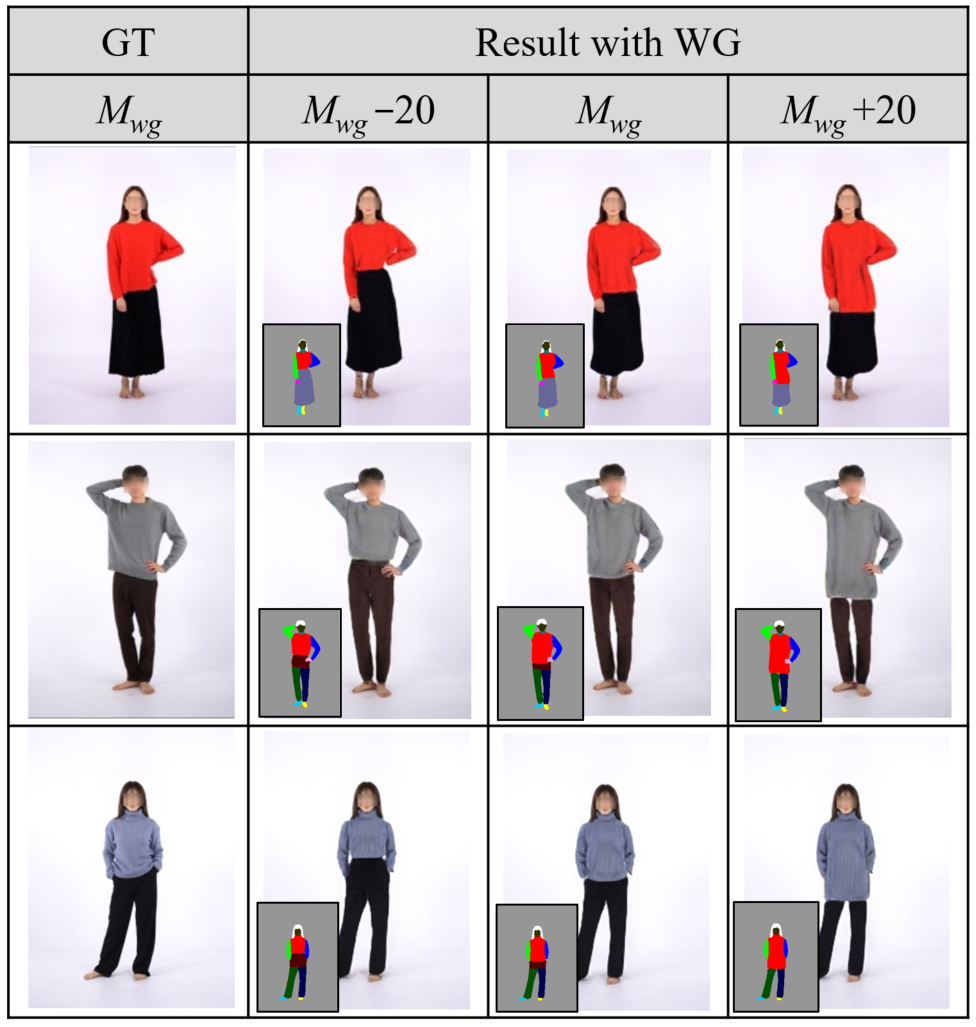}
	\caption{Synthesized results by controlling Wearing-guide Mask ($M_{wg}$). \textit{GT} indicate the ground-truth image and $M_{wg}$ means Wearing-guide Mask from the paired test set ($T_{pair}$). The 2nd and 4th column of the table show the results when using smaller and larger $M_{wg}$. The picture in picture images visualize the estimated parsing map $\hat{S}_h$ by WGPGM.}
	\label{fig:result_wgpgm}
\end{figure}

In this paper, we propose a method called \textit{WG-VITON} which generates virtual try-on images using the top and bottom clothes simultaneously with various wearing styles. Wearing-Guide Parsing Generation Module (i.e., Sec.\ref{sec:TB-WGPGM}) estimates a segmentation map of model with various wearing styles by considering the given combination of the clothes and \textit{Wearing-guide Mask}. Then following Structure-aware Clothes Warping Module (i.e., in Sec.\ref{sec:TB-SCWM}) warps images of the given clothes depending on the estimated segmentation map. Lastly, Try-On Module (i.e., in Sec.\ref{sec:TB-TOM}) uses the estimated segmentation map and the warped clothes to synthesize a realistic image of the model. The result of the experiment shows WG-VITON effectively synthesizes model's images considering the given top and bottom clothes simultaneously. In addition, the proposed wearing-guide scheme can generate images with various wearing styles as Fig.\ref{fig:result_wgpgm} illustrates, and also simulate complicated wearing styles such as partially tucking in the top to the bottom like Fig.\ref{fig:result_tuckin}.

\begin{figure}[t!]
	\centering
	\includegraphics[width=8cm]{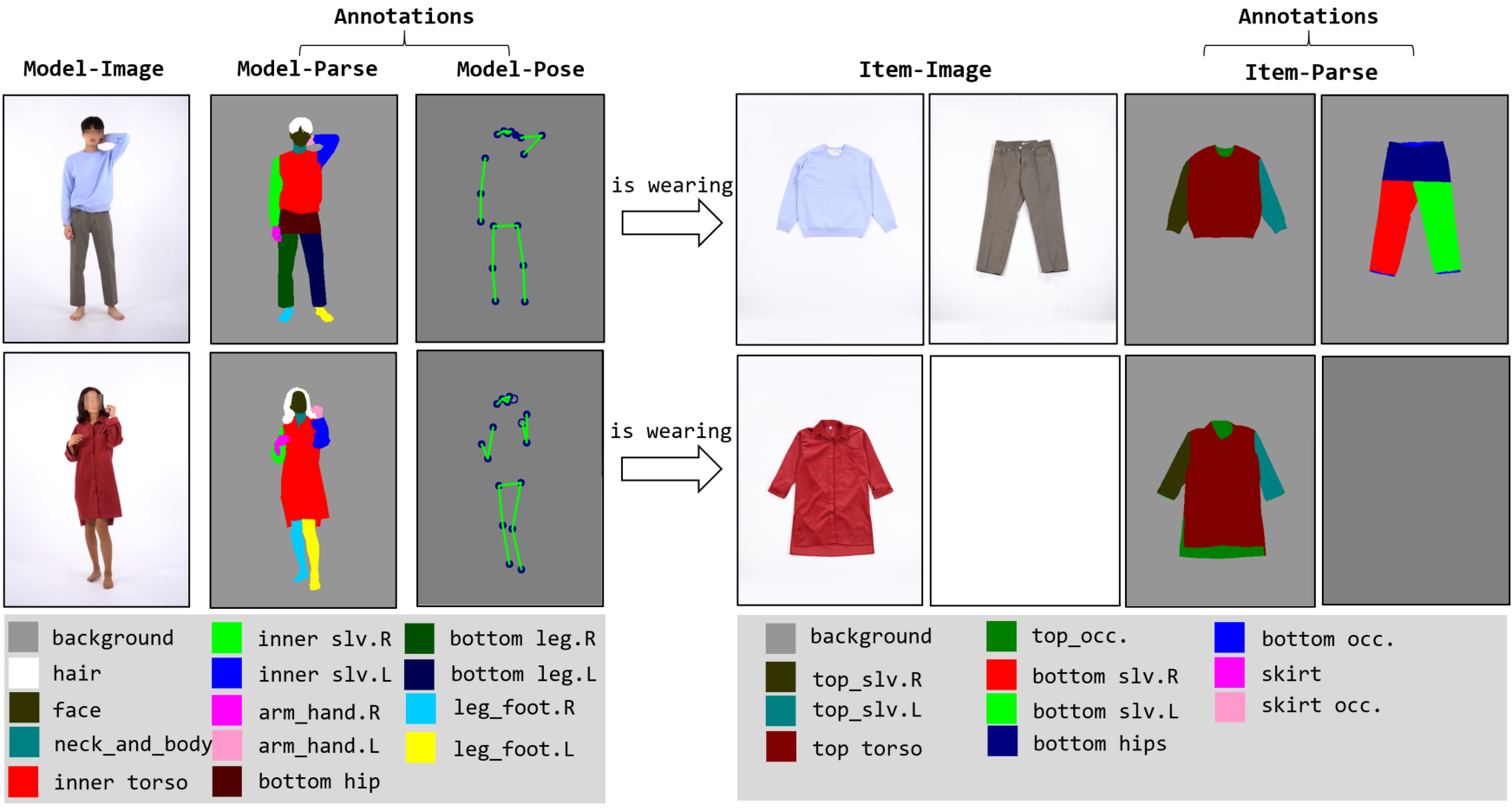}
	\caption{Data composition of FashionTB. Model's image and clothes' image are mapped by wearing relationship and each models and clothes have manual annotations. When a model wears a dress without the bottom, the mapping information of the bottom item will be ``None''.}
	\label{fig:fashiontb}
\end{figure}

\begin{figure}[t]
	\centering
	\includegraphics[width=7cm]{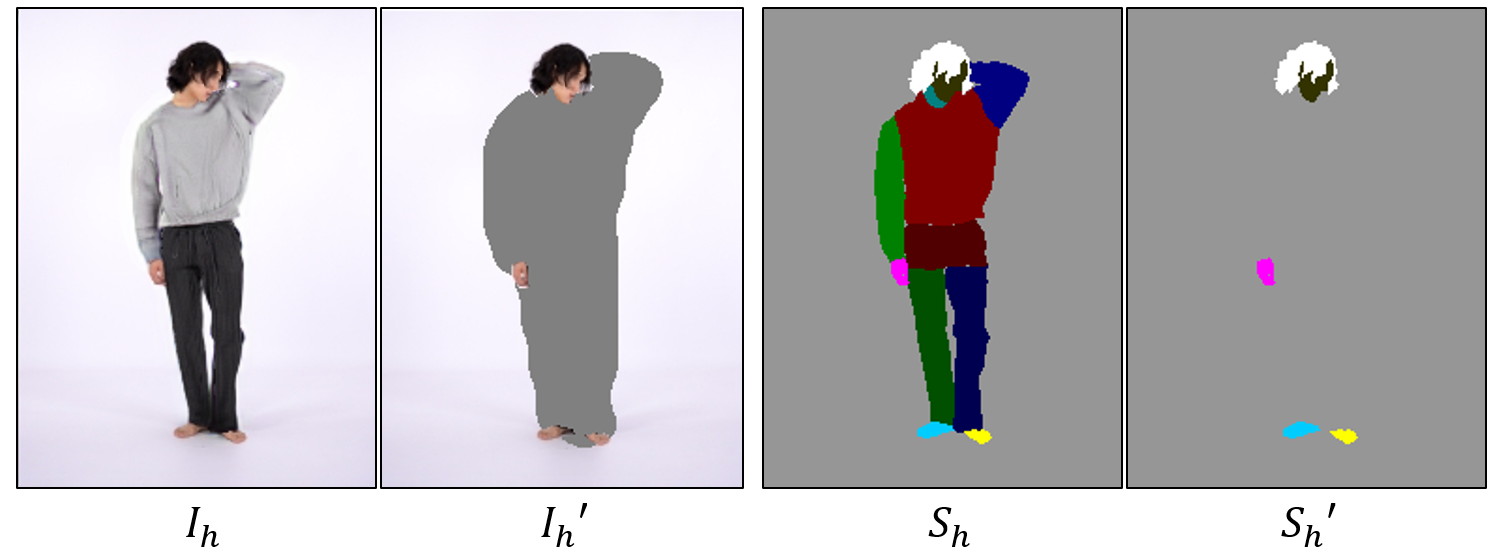}
	\caption{Data pre-processing for wearing-agnostic input.}
	\label{fig:agnostic}
\end{figure}

\section{Wearing-Guide Virtual Try-On}
We firstly explain a composition of the dataset because a composition of the dataset in the paper is differ from available VITON dataset such as \cite{han2018viton,dong2019MPV}. Sec.\ref{sec:TB-data} formulates data and introduces a data pre-processing method for constructing wearing-agnostic input. Then, Sec.\ref{sec:TB-WGPGM}--Sec.\ref{sec:TB-TOM} introduce the overall architecture of WG-VITON and how to train the network.

\subsection{Data and Pre-processing}
\label{sec:TB-data}
We construct a dataset of VITON for the top and bottom clothes by constructing sub-dataset of \cite{Park2022Data}. We name the dataset as \textit{FashionTB}. FashionTB provides images of models, tops, and bottoms and they are defined by $I_h \in \mathbb{R}^{3\times H \times W}$, $I_{top} \in \mathbb{R}^{3\times H \times W}$, and $I_{bt} \in \mathbb{R}^{3\times H \times W}$, respectively. Differ from other public dataset for VITON \cite{han2018viton, dong2019MPV}, each model and product have manual annotations. For the model data, we have human pose map $P_h\in \mathbb{R}^{17 \times H \times W}$ and segmentation $S_h \in \mathbb{R}^{17 \times H \times W}$. The top and bottom products have their own segmentation labels for their sleeves and torso (legs and hips for the bottom) $S_{top} \in \mathbb{R}^{3\times H \times W}$ and $S_{bt} \in \mathbb{R}^{3\times H \times W}$, respectively. Fig.\ref{fig:fashiontb} is a visualization of FashionTB and our research is performed under the dataset.

We construct two test set. One is $T_{pair}$ which has its own ground-truth image and labels, and the other is $T_{unpair}$ whose wearing infomation is ramdoly mixed to test arbitrary combination of the model and the products. We alleviate a dependency of input and training target by extending the pre-processing method in \cite{choi2021vitonhd}. Specifically, we eliminate the area of clothes in input model's image while remaining the area of model's hair, face, hands, and feet. Fig.\ref{fig:agnostic} illustrates the altered input $I_h'$ and $S_h'$, and the images do not have clues for what the model originally wears. 


\begin{figure}[t!]
	\includegraphics[width=8cm]{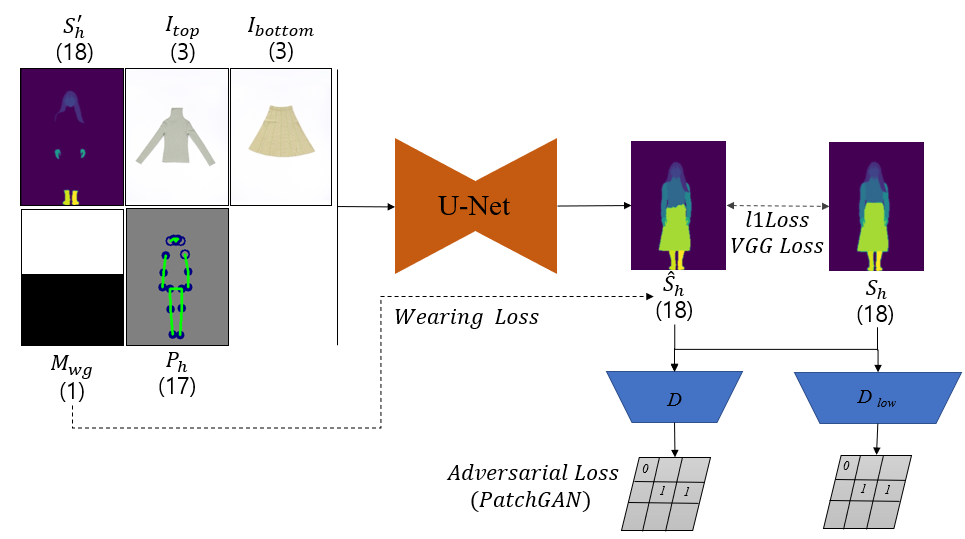}
	\caption{Architecture of WGPGM. The number below each symbol indicates the number of channels.}
	\label{fig:wgpgm_and_scwm}
\end{figure}


\subsection{Wearing-Guide Parsing Generation Module}
\label{sec:TB-WGPGM}
Prior VITON methods only for the top item generally maintained the bottom area and it provides clues for how to wear a given top product\cite{jandial2020sievenet,yang2020ACGPN,ge2021parserfree,choi2021vitonhd}. When a model of $I_h$ tuck in the top to the bottom, the final result of the existing studies follows the way of wearing. However, the wearing agnostic input of WG-VITON inevitably eliminates information to infer how to wear the top and bottom clothes, as Fig.\ref{fig:agnostic} describes. In this circumstance, the machine becomes confused because multiple answers exist for the identical input (i.e., combination of model and products). Thus, we propose \textit{Wearing-Guide Parsing Generation Module} (WGPGM) to alleviate the aforementioned problem by designing \textit{Wearing-guide Mask} and \textit{Wearing-guide Loss}.

Wearing-guide Mask $M_{wg}$ is a binary mask indicating the region where the bottom should not violate in the result of the parsing map. In the training phase, each pixel $(x,y)$ of $M_{wg}$ is assigned as

\begin{equation}
    M_{wg}(x,y)=\begin{cases}
            1, & y \leq maxy(S_{h_{(top,torso)}})
            \\
            0, & \text{otherwise}.
        \end{cases}
\end{equation}

\noindent 
where $maxy$ is a function to estimate the maximum y-coordinates of valid pixels and $S_{h_{(top,torso)}}$ is the segmentation map for torso of the top. 

Now, a function WGPGM can be formulated by

\begin{equation}
     \hat{S}_h = f_{WGPGM}(S_h', P_h, I_{top}, I_{bt}, M_{wg})
     \label{eq:fnpgm}
\end{equation}

\noindent
where $P_h$ is a set of human keypoints.

Fig.\ref{fig:wgpgm_and_scwm} (a) illustrates an overview architecture of WGPGM. We employ conditional GAN with U-Net\cite{ronneberger2015unet} and two PatchGAN discriminators \cite{isola2017patchgan}. While the first discriminator evaluates the entire generating parsing map, the second discriminator judges only the result of the lower half of the body (i.e., regions for a bottom item, legs, and feet.). The discriminator for the lower body part leverages the generation performance of the lower body part where FashionTB dataset has a larger variation. For example, the low body part in the dataset could be trousers, short pants, long/short skirt, or even empty when the model wears a long dress.

The training loss of the generator of WGPGM consists of cross-entropy loss, adversarial losses using LS-GAN\cite{mao2017lsgan}, feature matching losses\cite{wang2018high} and Wearing-guide Loss (Eq. \ref{eq:wearingloss}), and it can be formulated as
\begin{equation}
      \mathcal{L}_{WGPGM} = \lambda_{CE} \mathcal{L}_{CE} + \lambda_{adv} \mathcal{L}_{adv}+ \lambda_{FM} \mathcal{L}_{FM} + \lambda_{wg} \mathcal{L}_{wg}
     \label{eq:losspgm}
\end{equation}
\begin{equation}
      \mathcal{L}_{adv} = \mathcal{L}_{adv,D}+ \mathcal{L}_{adv,D_{low}}
     \label{eq:dloss}
\end{equation}
\begin{equation}
      \mathcal{L}_{FM} = \mathcal{L}_{FM,D} + \mathcal{L}_{FM,D_{low}}
     \label{eq:dbtloss}
\end{equation}
\begin{equation}
     \mathcal{L}_{wg} = |0 - M_{wg} \odot \hat{S}_{h_{bt}}|
     \label{eq:wearingloss}
\end{equation}

\noindent 
where all $\lambda$s are weights for the training and $\hat{S}_{h_{bt}}$ indicates regions for the bottom product.


\subsection{Structure-aware Clothes Warping Module}
\label{sec:TB-SCWM}
Structure-aware Clothes Warping Module (SCWM) utilizes Thin-Plate Spline (TPS) transformation \cite{rocco2017convolutional,han2018viton,choi2021vitonhd} and we extend it to simultaneously warp the top and bottom products. SCWM is a function $f_{SCWM}$ to estimate warping parameters for the given garments from the estimated segmentation in WGPGM, human pose, and an image of the top and bottom such as Eq.\ref{eq:scwm}.

\begin{equation}
     \theta_{top}, \theta_{bt} = f_{SCWM}(\hat{S}_{h_{top}}, \hat{S}_{h_{bt}}, P_h, I_{top}, I_{bt})
     \label{eq:scwm}
\end{equation}

\noindent 
where $\theta_{top}$ and $\theta_{bt}$ are parameters for the TPS transform for the top and bottom products, respectively. 

The model's information (i.e., $P_h$, $\hat{S}_{h_{top}}$, and $\hat{S}_{h_{bt}}$) and product images (i.e., $I_{top}$ and $I_{bt}$) are analyzed by each set of convolution layers, and the top and bottom clothes sequentially go through the identical convolution layers rather than concatenating them. We perform correlation matching twice with the estimated features, and then apply the TPS transform using the estimated parameters $\theta_{top}$ and $\theta_{bt}$.  
In contrast to existing studies which train the warping using L1 loss only for color, we add L1 losses for semantic segmentation (3-channel masks) of the clothes. The approach has two main advantages: (i) SCWM is properly trained by considering not only the color, but also structure of the clothes distinguishing the area of torso and sleeves or hip and legs and (ii) SCWM is still able to warp the items when the colors of target clothes and background are similar (e.g., white clothes with a white background). 




\subsection{Try-On Module}
\label{sec:TB-TOM}
Try-On Module (TOM) finally synthesizes a model image using the estimation results from the previous modules. Specifically, $\hat{S}_h$ from WGPGM, $\hat{I}_{top}$ and $\hat{I}_{bt}$ from SCWM, and $I'_h$ and $P_h$ from data are the inputs of TOM. Then, the function TOM can be formulated by 

\begin{equation}
     \hat{I}_h = f_{TOM}(I_h', P_h, \hat{S}_h, \hat{I}_{top}, \hat{I}_{bt})
     \label{eq:tom}
\end{equation}

We use U-Net based generator and the mask composition \cite{ronneberger2015unet,han2018viton,wang2018cpvton}. Specifically, U-Net estimates two binary masks for the composition (i.e., $\hat{M}_{top}$ and $\hat{M}_{bt}$), and a base of the synthesized image (i.e., $\hat{I}_{base}$). We synthesize the final result by compositing $I_h'$, $\hat{I}_{top}$, $\hat{I}_{bt}$, and $\hat{I}_h$ using the masks $M_{I_h'}$, $\hat{M}_{top}$ and $\hat{M}_{bt}$. Specifically, three mask compositions are sequentially performed to make the result of the synthesis $\hat{I}$.

\begin{figure*}[t]
	\centering
	\includegraphics[width=13.5cm]{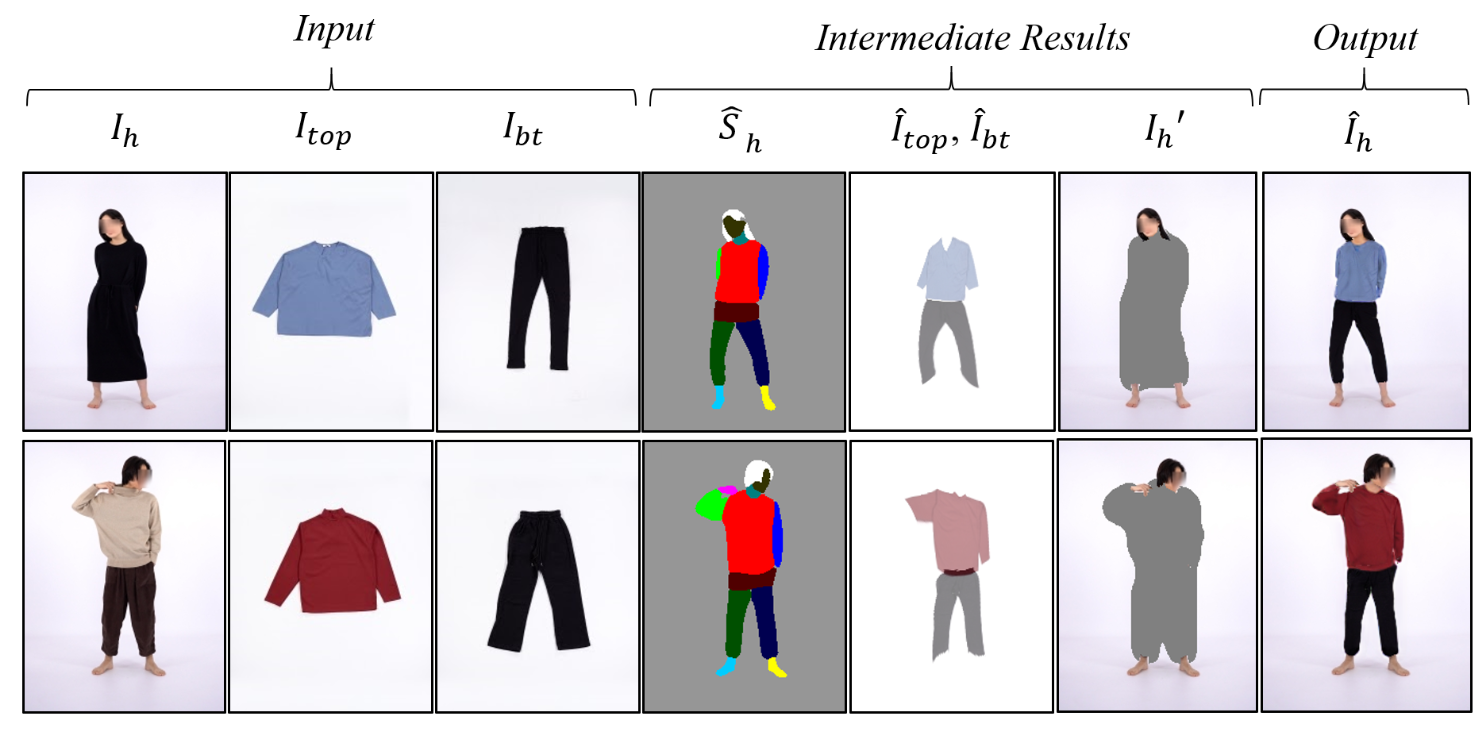}
	\caption{The result of WG-VITON in $T_{unpair}$ with intermediate estimation results of each modules. For $\hat{I}_{top}$ and $\hat{I}_{bt}$, we overlap two transparent images. Images in the last column indicate the final result of WG-VITON.}
	\label{fig:result_vitontb}
\end{figure*}

\section{Experiments}
We perform quantitative and qualitative evaluations for WG-VITON. All experiments in this work utilize FashionTB dataset mentioned in Sec.\ref{sec:TB-data} because no public dataset contain mapping between the tops, bottoms, and models.

\subsection{Evaluation}
\label{sec:results}



\noindent{\textbf{Qualitative Evaluation}} Fig.\ref{fig:result_vitontb} illustrates the final result of the proposed method and its intermediate estimations. Images in the 1st--3rd columns are inputs for WG-VITON and we omit $S_h$ and $P_h$ of each sample. The 4th column depicts a result of WGPGM. WGPGM effectively estimates segmentation regions considering the given model and clothes. Following SCWM applies TPS transform to $I_{top}$ and $I_{bt}$ by referring $\hat{S}_h$ from WGPGM, and warps images of the clothes $\hat{I}_{top}$ and $\hat{I}_{bt}$ (overlapped image in the 5th column). With the estimation results and wearing-agnostic human image $I_h'$ (the 6th column), TOM synthesizes the model image wearing given clothes while maintaining other characteristics of model such as identity and pose (the last column).

\noindent{\textbf{Quantitative Evaluation}} We use Structural Similarity (SSIM)\cite{wang2004SSIM}, Fr\'echet Inception Distance (FID)\cite{heusel2017FID}, and Learned Perceptual Image Patch Similarity (LPIPS)\cite{zhang2018LPIPS} to evaluate the baseline network. SSIM and LPIPS are applied to $T_{pair}$ which has the ground-truth image for wearing pairs, and FID is used for both $T_{pair}$ and $T_{unpair}$. Table \ref{tb:quantitative_result_ideal} shows results under the four evaluation cases. 

\begin{table}[t!]
\centering
\begin{tabular}{|c||c|c|c|c|}
\hline
\begin{tabular}[c]{@{}l@{}}Resol.\\ \end{tabular} & \begin{tabular}[c]{@{}c@{}}SSIM$\uparrow$\\ ($T_{pair}$)\end{tabular} & \begin{tabular}[c]{@{}c@{}}LPIPS$\downarrow$\\ ($T_{pair}$)\end{tabular} & \begin{tabular}[c]{@{}c@{}}FID$\downarrow$\\ ($T_{pair}$)\end{tabular} &
\begin{tabular}[c]{@{}c@{}}FID$\downarrow$\\ ($T_{unpair}$)\end{tabular}\\ \hline\hline
256$\times$192                                                   & 0.901                                                       &         0.065              & 10.184 &    12.663             \\ \cline{1-5}
512$\times$384                                                  & 0.911                                                       & 0.069                                                        &  12.991                    & 16.359                                    \\ \hline
\end{tabular}
\caption{Qualitative evaluation of WG-VITON. Up arrow in the table indicates that the higher value is the better for the evaluation metric. We mention a related test dataset below the name of each metric.}
\label{tb:quantitative_result_ideal}
\end{table}

\subsection{Style Generation using WGPGM}
\label{sec:result_WGPGM}
\begin{figure}[t!]
	\centering
	\includegraphics[width=7.5cm]{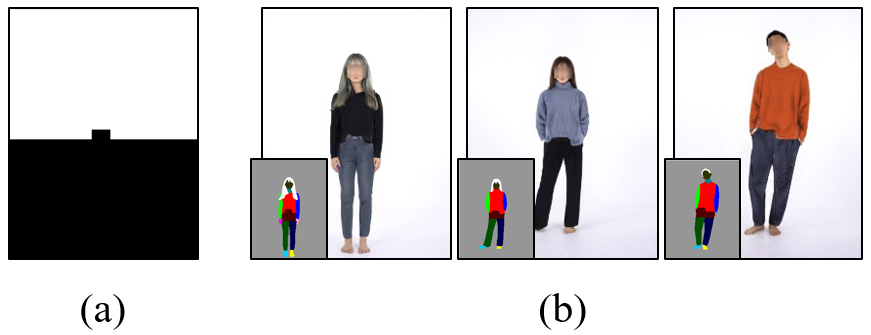}
	\caption{(a): Example of the Wearing-guide Mask  and (b): Results by the mask where each model partly tucks in the top to the bottom.}
	\label{fig:result_tuckin}
\end{figure}

Verifying the effectiveness of the wearing-guide scheme introduced in Sec.\ref{sec:TB-WGPGM}, we synthesize $T_{pair}$ samples with different $M_{wg}$ as Fig.\ref{fig:result_wgpgm} shows. The 1st column of the figure shows the ground-truth image of $T_{pair}$. Without the wearing-guide scheme, parsing map generator will estimate a map that minimizes the training loss among various wearing styles in the training set. On the other hand, as the 2nd--4th columns illustrate, WG-VITON can simulate various wearing styles by controlling $M_{wg}$. Specifically, the results in the 3rd column use the ground-truth $M_{wg}$ so their styles are similar to those of the ground-truth. The 2nd and 4th columns show the results when we decrease and increase $M_{wg}$ by 20 pixels, respectively. As a result, images with smaller $M_{wg}$ tend to enlarge an area of the bottom item while images with higher $M_{wg}$ expand the top to cover the hips of the model. In addition, when we use a relatively complicated mask like Fig.\ref{fig:result_tuckin}(a), WG-VITON can synthesize an image where a model wears the top tucked in only a part into the bottom, which is one of the trendy wearing styles as Fig.\ref{fig:result_tuckin}(b) illustrates.

\section{Conclusion}
In this paper, we propose WG-VITON which synthesizes the model's image wearing the target top and bottom clothes with various wearing styles. Using the given clothes and model, WG-VITON can generate parsing map with various wearing styles, warp the images of the garments according to the target model, and make the model's image using the results of the previous modules. In specific, in Wearing-Guide Parsing Generation Module (i.e., WGPGM), we can control a length of the top clothes in output image controlling Wearing-guide Mask. Moreover, when the Mask having complicated shape, we can simulate wearing styles such as tucking in a part of the top to the bottom clothes. We believe that WG-VITON provides interesting insights of simulating wearing styles in fashion and lead the following researches to make VITON technology more applicable.

{\small
\bibliographystyle{ieee_fullname}
\bibliography{egbib}

\begin{thebibliography}{10}\itemsep=-1pt

\bibitem{choi2021vitonhd}
Seunghwan Choi, Sunghyun Park, Minsoo Lee, and Jaegul Choo.
\newblock Viton-hd: High-resolution virtual try-on via misalignment-aware
  normalization.
\newblock In {\em Proceedings of the IEEE/CVF Conference on Computer Vision and
  Pattern Recognition}, pages 14131--14140, 2021.

\bibitem{Cui_2021_Dior}
Aiyu Cui, Daniel McKee, and Svetlana Lazebnik.
\newblock Dressing in order: Recurrent person image generation for pose
  transfer, virtual try-on and outfit editing.
\newblock In {\em Proceedings of the IEEE/CVF International Conference on
  Computer Vision (ICCV)}, pages 14638--14647, October 2021.

\bibitem{dong2019MPV}
Haoye Dong, Xiaodan Liang, Xiaohui Shen, Bochao Wang, Hanjiang Lai, Jia Zhu,
  Zhiting Hu, and Jian Yin.
\newblock Towards multi-pose guided virtual try-on network.
\newblock In {\em Proceedings of the IEEE/CVF International Conference on
  Computer Vision}, pages 9026--9035, 2019.

\bibitem{ge2021disentangled}
Chongjian Ge, Yibing Song, Yuying Ge, Han Yang, Wei Liu, and Ping Luo.
\newblock Disentangled cycle consistency for highly-realistic virtual try-on.
\newblock In {\em Proceedings of the IEEE/CVF Conference on Computer Vision and
  Pattern Recognition}, pages 16928--16937, 2021.

\bibitem{ge2021parserfree}
Yuying Ge, Yibing Song, Ruimao Zhang, Chongjian Ge, Wei Liu, and Ping Luo.
\newblock Parser-free virtual try-on via distilling appearance flows.
\newblock In {\em Proceedings of the IEEE/CVF Conference on Computer Vision and
  Pattern Recognition}, pages 8485--8493, 2021.

\bibitem{han2018viton}
Xintong Han, Zuxuan Wu, Zhe Wu, Ruichi Yu, and Larry~S Davis.
\newblock Viton: An image-based virtual try-on network.
\newblock In {\em Proceedings of the IEEE conference on computer vision and
  pattern recognition}, pages 7543--7552, 2018.

\bibitem{heusel2017FID}
Martin Heusel, Hubert Ramsauer, Thomas Unterthiner, Bernhard Nessler, and Sepp
  Hochreiter.
\newblock Gans trained by a two time-scale update rule converge to a local nash
  equilibrium.
\newblock {\em Advances in neural information processing systems}, 30, 2017.

\bibitem{isola2017patchgan}
Phillip Isola, Jun-Yan Zhu, Tinghui Zhou, and Alexei~A Efros.
\newblock Image-to-image translation with conditional adversarial networks.
\newblock In {\em Proceedings of the IEEE conference on computer vision and
  pattern recognition}, pages 1125--1134, 2017.

\bibitem{jandial2020sievenet}
Surgan Jandial, Ayush Chopra, Kumar Ayush, Mayur Hemani, Balaji Krishnamurthy,
  and Abhijeet Halwai.
\newblock Sievenet: A unified framework for robust image-based virtual try-on.
\newblock In {\em Proceedings of the IEEE/CVF Winter Conference on Applications
  of Computer Vision}, pages 2182--2190, 2020.

\bibitem{li2021ovnet}
Kedan Li, Min~Jin Chong, Jeffrey Zhang, and Jingen Liu.
\newblock Toward accurate and realistic outfits visualization with attention to
  details.
\newblock In {\em Proceedings of the IEEE/CVF Conference on Computer Vision and
  Pattern Recognition}, pages 15546--15555, 2021.

\bibitem{mao2017lsgan}
Xudong Mao, Qing Li, Haoran Xie, Raymond~YK Lau, Zhen Wang, and Stephen
  Paul~Smolley.
\newblock Least squares generative adversarial networks.
\newblock In {\em Proceedings of the IEEE international conference on computer
  vision}, pages 2794--2802, 2017.

\bibitem{minar2020cpvtonplus}
Matiur~Rahman Minar, Thai~Thanh Tuan, Heejune Ahn, Paul Rosin, and Yu-Kun Lai.
\newblock Cp-vton+: Clothing shape and texture preserving image-based virtual
  try-on.
\newblock In {\em CVPR Workshops}, 2020.

\bibitem{neuberger2020oviton}
Assaf Neuberger, Eran Borenstein, Bar Hilleli, Eduard Oks, and Sharon Alpert.
\newblock Image based virtual try-on network from unpaired data.
\newblock In {\em Proceedings of the IEEE/CVF Conference on Computer Vision and
  Pattern Recognition}, pages 5184--5193, 2020.

\bibitem{Park2022Data}
Soonchan Park, Hanbyeol Yoo, Johan Lee, and Jiyoung Park.
\newblock High resolution dataset for virtual try-on utilizing multiple
  products and its application.
\newblock {\em Korea Transactions on Computing Practices}, 28(1):68--73, 2022.

\bibitem{rocco2017convolutional}
Ignacio Rocco, Relja Arandjelovic, and Josef Sivic.
\newblock Convolutional neural network architecture for geometric matching.
\newblock In {\em Proceedings of the IEEE conference on computer vision and
  pattern recognition}, pages 6148--6157, 2017.

\bibitem{ronneberger2015unet}
Olaf Ronneberger, Philipp Fischer, and Thomas Brox.
\newblock U-net: Convolutional networks for biomedical image segmentation.
\newblock In {\em International Conference on Medical image computing and
  computer-assisted intervention}, pages 234--241. Springer, 2015.

\bibitem{wang2018cpvton}
Bochao Wang, Huabin Zheng, Xiaodan Liang, Yimin Chen, Liang Lin, and Meng Yang.
\newblock Toward characteristic-preserving image-based virtual try-on network.
\newblock In {\em Proceedings of the European Conference on Computer Vision
  (ECCV)}, pages 589--604, 2018.

\bibitem{wang2018high}
Ting-Chun Wang, Ming-Yu Liu, Jun-Yan Zhu, Andrew Tao, Jan Kautz, and Bryan
  Catanzaro.
\newblock High-resolution image synthesis and semantic manipulation with
  conditional gans.
\newblock In {\em Proceedings of the IEEE conference on computer vision and
  pattern recognition}, pages 8798--8807, 2018.

\bibitem{wang2004SSIM}
Zhou Wang, Alan~C Bovik, Hamid~R Sheikh, and Eero~P Simoncelli.
\newblock Image quality assessment: from error visibility to structural
  similarity.
\newblock {\em IEEE transactions on image processing}, 13(4):600--612, 2004.

\bibitem{yang2020ACGPN}
Han Yang, Ruimao Zhang, Xiaobao Guo, Wei Liu, Wangmeng Zuo, and Ping Luo.
\newblock Towards photo-realistic virtual try-on by adaptively
  generating-preserving image content.
\newblock In {\em Proceedings of the IEEE/CVF Conference on Computer Vision and
  Pattern Recognition}, pages 7850--7859, 2020.

\bibitem{zhang2018LPIPS}
Richard Zhang, Phillip Isola, Alexei~A Efros, Eli Shechtman, and Oliver Wang.
\newblock The unreasonable effectiveness of deep features as a perceptual
  metric.
\newblock In {\em Proceedings of the IEEE conference on computer vision and
  pattern recognition}, pages 586--595, 2018.

\end{thebibliography}
}

\end{document}